\documentclass[sigconf]{acmart}

\usepackage{multirow}
\usepackage{booktabs}
\usepackage{makecell}
\usepackage{array} 
\usepackage{balance}

\newcommand{\etal}{\textit{et al}.}
\newcommand{\ie}{\textit{i}.\textit{e}.}
\newcommand{\eg}{\textit{e}.\textit{g}.}
\newcommand{\etc}{\textit{etc}.}

\AtBeginDocument{%
  \providecommand\BibTeX{{%
    \normalfont B\kern-0.5em{\scshape i\kern-0.25em b}\kern-0.8em\TeX}}}

\copyrightyear{2023} 
\acmYear{2023} 
\setcopyright{acmlicensed}\acmConference[MM '23]{Proceedings of the 31st ACM International Conference on Multimedia}{October 29-November 3, 2023}{Ottawa, ON, Canada}
\acmBooktitle{Proceedings of the 31st ACM International Conference on Multimedia (MM '23), October 29-November 3, 2023, Ottawa, ON, Canada}
\acmPrice{15.00}
\acmDOI{10.1145/3581783.3612482}
\acmISBN{979-8-4007-0108-5/23/10}

\acmSubmissionID{3417}

\begin{document}

\title{SDDNet: Style-guided Dual-layer Disentanglement Network for Shadow Detection}

\author{Runmin Cong}
\authornote{Runmin Cong and Wei Zhang are also affiliated with the Key Laboratory of Machine Intelligence and System Control, Ministry of Education, Jinan, Shandong, China.}
\email{rmcong@sdu.edu.cn}
\affiliation{%
  \institution{Shandong University}
  \city{Jinan}
  \state{Shandong}
  \country{China}
}

\author{Yuchen Guan}
\email{gyc23@mails.tsinghua.edu.cn}
\affiliation{%
  \institution{Beijing Jiaotong University}
  \city{Beijing}
  \country{China}
}

\author{Jinpeng Chen}
\authornote{Corresponding author.}
\email{jinpechen2-c@my.cityu.edu.hk}
\affiliation{%
  \institution{City University of Hong Kong}
  \city{Hong Kong SAR}
  \country{China}
}

\author{Wei Zhang}
\authornotemark[1]
\email{davidzhang@sdu.edu.cn}
\affiliation{%
  \institution{Shandong University}
  \city{Jinan}
  \state{Shandong}
  \country{China}
}

\author{Yao Zhao}
\email{yzhao@bjtu.edu.cn}
\affiliation{%
  \institution{Beijing Jiaotong University}
  \city{Beijing}
  \country{China}
}

\author{Sam Kwong}
\email{cssamk@cityu.edu.hk}
\affiliation{%
  \institution{City University of Hong Kong \& Lingnan University}
  \city{Hong Kong SAR}
  \country{China}
}

\renewcommand{\shortauthors}{Runmin Cong, et al.}

\begin{abstract}
Despite significant progress in shadow detection, current methods still struggle with the adverse impact of background color, which may lead to errors when shadows are present on complex backgrounds. Drawing inspiration from the human visual system, we treat the input shadow image as a composition of a background layer and a shadow layer, and design a Style-guided Dual-layer Disentanglement Network (SDDNet) to model these layers independently. To achieve this, we devise a Feature Separation and Recombination (FSR) module that decomposes multi-level features into shadow-related and background-related components by offering specialized supervision for each component, while preserving information integrity and avoiding redundancy through the reconstruction constraint. Moreover, we propose a Shadow Style Filter (SSF) module to guide the feature disentanglement by focusing on style differentiation and uniformization. With these two modules and our overall pipeline, our model effectively minimizes the detrimental effects of background color, yielding superior performance on three public datasets with a real-time inference speed of 32 FPS. Our code is publicly available at: \textit{\url{https://github.com/rmcong/SDDNet_ACMMM23}}.
\end{abstract}





\maketitle

\section{Introduction}
Shadows are a ubiquitous illumination phenomenon resulting from the linear propagation of light, which can negatively affect tasks such as object detection \cite{crm/tip21/DAFNet,crm/tgrs22/RRNet}. Accurate shadow detection can provide valuable insights into scene geometry \cite{okabe2009attached, karsch2011rendering} and light source positioning \cite{lalonde2012estimating, panagopoulos2009robust}, thereby enhancing scene understanding. 
As a result, shadow detection has become a foundational task in computer vision, attracting increasing attention in recent years.

\begin{figure}[t]
\centering
\includegraphics[width=0.48\textwidth]{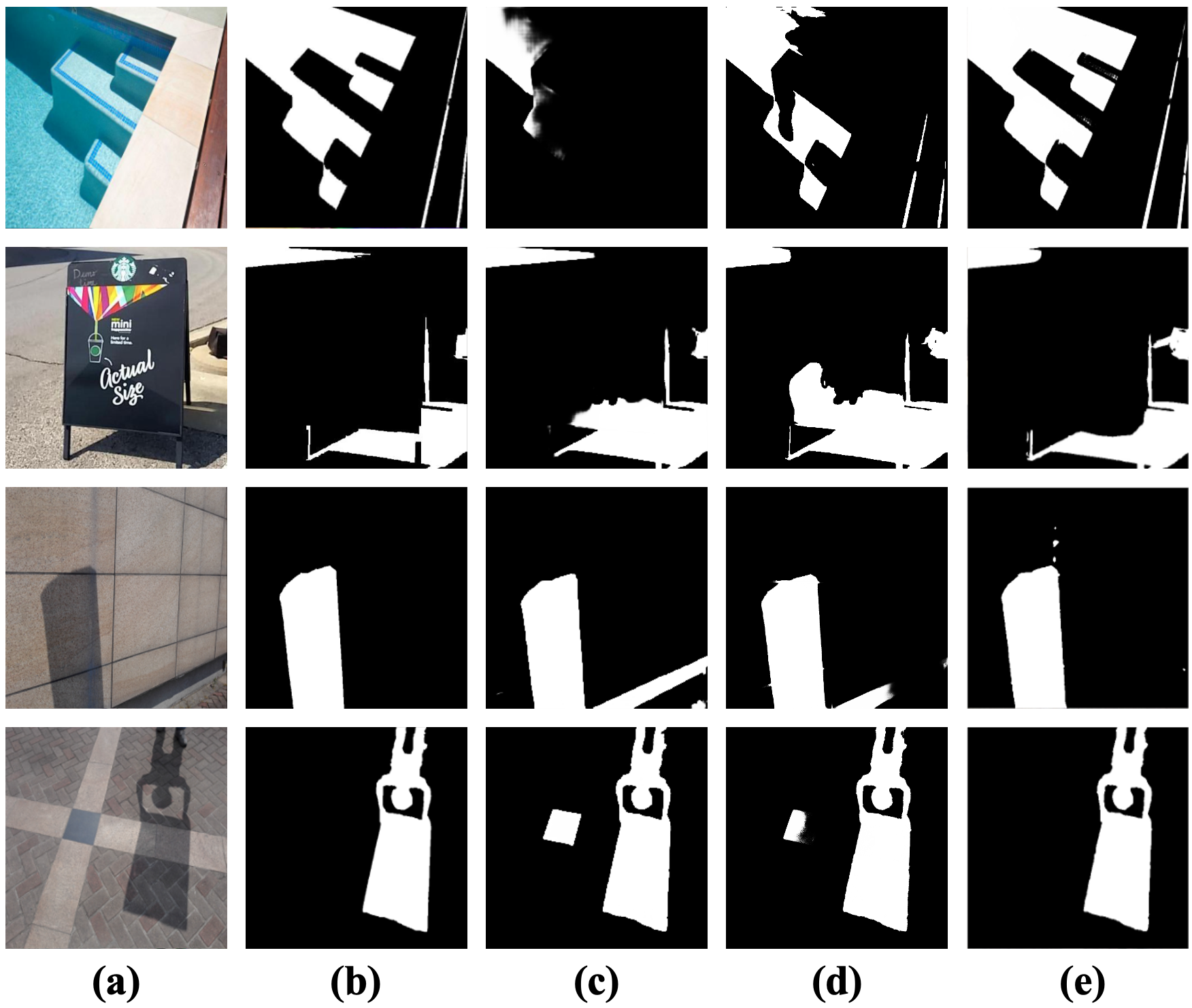}
\caption{Some difficult cases in shadow detection. (a) The input images. (b) The ground truth shadow maps. (c) The predicted results of ECA \cite{fang2021robust}. (d) The predicted results of MTMT-Net \cite{chen2020multi}. (e) The predicted results of our SDDNet.}
\label{visual1}
\end{figure}

Early shadow detection efforts mainly relied on physical methods, constructing physical illumination models \cite{guo2012paired, guo2011single} and employing hand-crafted features, \eg, illumination cues \cite{finlayson2009entropy, panagopoulos2011illumination} and texture \cite{zhu2010learning, vicente2017leave}. However, with the remarkable performance of convolutional neural networks (CNNs) \cite{cong2022cir, cong2022does, cong2022global, cong2022weakly, cong2023multi, cong2023pugan,crm/tetci22/PSNet,crm/tcyb21/ASIFNet,crm/tip21/DPANet,crm/tip/WaveNet,crm/eccv20/RGBDSOD,crm/jbhi22/polyp,crm/tim22/covid,crm/acmmm21/CDINet, zhang2023robust, zhao2023bilateral,crm/tip20/MCMT-GAN,crm/spl21/underwater,crm/access17/dsr, zhang2023controlvideo} in computer vision tasks, deep learning-based methods \cite{zhu2021mitigating, zheng2019distraction, chen2020multi, liao2021shadow, chen2021triple, wang2020instance, fang2021robust, panagopoulos2009robust, hu2021revisiting} have progressively become the mainstream of shadow detection.
Previous deep learning-based shadow detection methods predominantly focused on guidance from location, semantics, and context perspectives, without attaching importance to the detrimental influence of background color. 
Consequently, these detectors tend to associate shadow features with dark colors, leading to incorrect detection results in some complex scenes. Specifically, errors can be divided into two categories: 1) weak shadow regions on light-colored backgrounds (\eg, the first example in Figure \ref{visual1}), which are wrongly detected as non-shadow regions, and 2) dark-colored background areas (\eg, the second example in Figure \ref{visual1}), which are often misclassified as shadows. In summary, detection results are greatly affected by background color, and similar shadows in different backgrounds may yield completely disparate detection outcomes. For instance, detecting a shadow is relatively simple when it forms on a ground composed entirely of light-colored bricks, but becomes challenging when the shadow appears on a ground with alternating dark and light bricks, as shown in the third and fourth examples of Figure \ref{visual1}. In these scenarios, the shadow on the light-colored brick is difficult to detect, while the dark brick is prone to being misidentified as a shadow, irrespective of whether it is in shadow or not.

Since the shadows are created by light being blocked, they are inherently colorless.
Based on this, humans can discern shadows on complex backgrounds through a three-step process: first, recognizing background attributes; second, identifying shadow attributes by observing confident shadow regions; and finally, detecting all shadow regions based on our understanding of background and shadow attributes. Inspired by this process, we propose treating shadow images as a composition of background and shadow layers, and modeling them separately to effectively reduce the impact of background color on detection performance. This objective can be accomplished through the strategy of feature disentanglement, which is proved to be effective in many computer vision works. For instance, \cite{yang2020task} conducted an early trial to incorporate task-feature co-disentanglement regularizations for multi-task learning and achieved satisfactory performance. 

In this paper, we introduce the concept of feature disentanglement into shadow detection to realize the separated modeling of background and shadow layers. We present a novel Style-guided Dual-layer Disentanglement Network (SDDNet) featuring two innovative modules, \ie, the Feature Separation and Recombination (FSR) module and the Shadow Style Filter (SSF) module. The FSR module effectively decomposes multi-level features into background-related and shadow-related components, which is explicitly achieved by providing distinct supervision for each set of components. The shadow-related component receives supervision from the ground-truth shadow map, while the background-related component is guided to generate a shadow-free background image. Furthermore, to ensure information integrity, the recombined features merged from both components are supervised to reconstruct the input image. During prediction, only the shadow-related component is utilized to generate the final shadow map, effectively eliminating the adverse influence of the background. Additionally, to further constrain the FSR module on feature disentanglement, particularly for background-related component that lacks a background ground-truth, we propose a Shadow Style Filter (SSF) module to extract and constrain style attributes of the shadow-related component, background-related component, and recombined features. Specifically, we regard the presence or absence of shadows as a style. From this perspective, the recombined features and shadow-related component should have consistent styles, while the background-related component should exhibit a different style from them. Based on this principle, we can generate background images in an indirect style-guided manner, thereby facilitating feature disentanglement within the FSR module. 

In summary, our contributions are primarily three-fold:
\begin{itemize}
\item We model the shadow image as a superposition of shadow layers on background layers, and then propose a Style-guided Dual-layer Disentanglement Network (SDDNet) for shadow detection. Extensive experiments on three public datasets demonstrate that our proposed method outperforms all state-of-the-art shadow detection methods with a real-time inference speed of 32 FPS.
\item We design a Feature Separation and Recombination (FSR) module to decompose image features into shadow-related and background-related components, thereby preventing predictions from being misled by background information.
\item We devise a Shadow Style Filter (SSF) module that assists feature separation through style differentiation and uniformization, especially to help generate the background-related component in an indirect style-guided manner.
\end{itemize}

\section{Related Works}

\subsection{Shadow Detection}
Related works on shadow detection can be broadly categorized into traditional methods and deep learning-based methods.

Early efforts in shadow detection primarily focused on constructing physical illumination models \cite{le2019shadow, finlayson2005removal, finlayson2009entropy, salvador2004cast, panagopoulos2011illumination, huang2011characterizes, guo2012paired, lalonde2010detecting, guo2011single} to analyze the shadow formation process. Based on these models, shadows were detected either by using physical models \cite{finlayson2005removal, finlayson2009entropy} or by employing traditional machine learning-based detectors with hand-crafted features, such as illumination cues \cite{finlayson2009entropy, panagopoulos2011illumination, guo2012paired, guo2011single}, texture \cite{guo2012paired, zhu2010learning}, and edge \cite{huang2011characterizes, lalonde2010detecting}. Although these methods led to improvements, most of them relied on assumptions (\eg, fixed background classes, uniform illumination, \etc) that are difficult to satisfy in complex situations. Additionally, the hand-crafted features may not be discriminative enough for detecting intricate shadow regions.

Inspired by the outstanding performance of CNN in computer vision tasks, deep learning-based methods \cite{zhu2021mitigating, zheng2019distraction, chen2020multi, liao2021shadow, zheng2019distraction, chen2021triple, wang2020instance, fang2021robust, panagopoulos2009robust, hu2021revisiting, lu2022video, lyu2022portrait, he2021unsupervised, hu2019direction} have gained popularity in shadow detection. With their ability to extract and select discriminative features, CNNs are more robust than traditional methods that use hand-crafted features. Khan \etal\ \cite{panagopoulos2009robust} were the first to apply CNNs to shadow detection, extracting features from superpixels using a 7-layer CNN and feeding these features to a conditional random field (CRF) model to refine the detection results. Zheng \etal\ \cite{zheng2019distraction} integrated the semantics of distraction regions to extend CNNs for robust shadow detection. Some researchers \cite{lyu2022portrait, he2021unsupervised, le2018a+, nguyen2017shadow} employed generative adversarial networks (GAN) \cite{goodfellow2020generative} for shadow detection.
Recently, Chen \etal\ \cite{chen2020multi} proposed a semi-supervised teacher-student framework to detect shadow regions, edges, and count under consistency constraints. Zhu \etal\ \cite{zhu2021mitigating} introduced a feature reweighting method to balance the intensity-variant and intensity-invariant features obtained by self-supervised decomposition. Liao \etal\ \cite{liao2021shadow} incorporated confidence maps into shadow detection and combined the prediction results of multiple methods for shadow detection. 

Despite the significant improvements offered by these methods, they still suffer from background color interference. This interference causes confusion between dark background areas and shadow regions, as well as between light background areas and weak shadow regions. In this study, we disentangle background-related and shadow-related components, utilizing only the shadow-related component to predict the final results. This approach enhances the robustness of shadow detection in complex scenes.

\begin{figure*}[!t]
    \centering
    \includegraphics[width=0.9\textwidth]{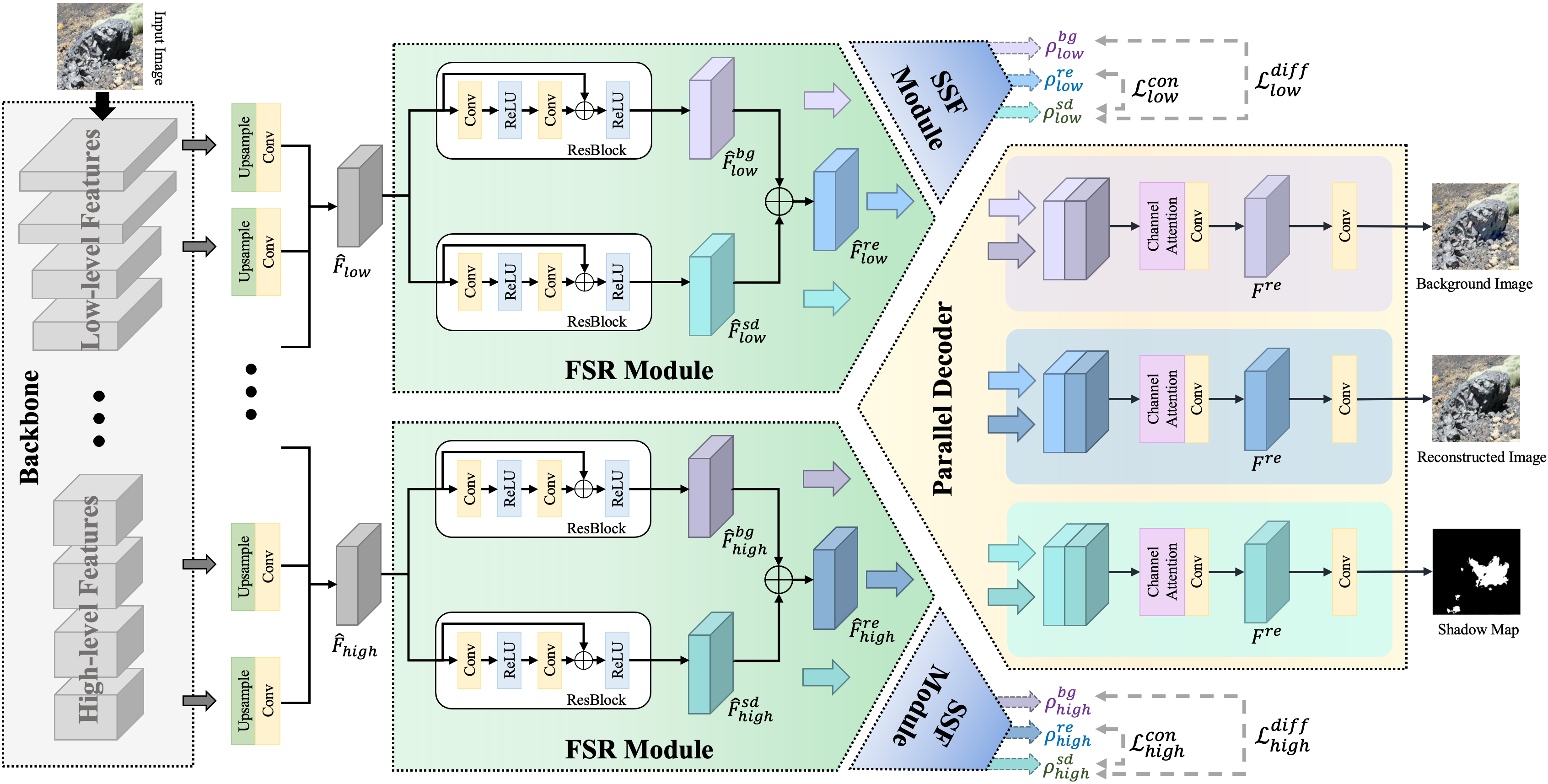}
    \caption{Architecture of the proposed SDDNet. Given an input image, SDDNet outputs the shadow map, background image, and reconstructed image in an end-to-end manner. Firstly, the backbone extracts integrated low-level and high-level features. Then, the proposed FSR module decomposes the features and produce shadow-related component, background-related component, and recombined features. In addition, the SSF module extracts style attributes and guide the feature disentanglement process. Finally, the low-level and high-level features are fused through the parallel decoder to generate three outputs (\ie, background image, shadow map, and reconstructed input image).}
    \label{pipeline}
\end{figure*}

\subsection{Style Transfer}
In the domain of neural style transfer, research has been conducted to comprehend the content and style of image features. Gatys \etal\ \cite{gatys2016image} proposed utilizing the Gram matrix of image features as a means to encapsulate the distinctive style of an image. Subsequent studies \cite{lee2022fifo, johnson2016perceptual} have further corroborated the efficacy of the Gram matrix in capturing and representing image styles.

In this paper, we employ the Gram matrix to extract style attributes and regulate the consistency or diversity of these attributes across various features and components of the input shadow image. This approach serves to bolster our feature disentanglement process, ultimately leading to enhanced outcomes.

\section{Proposed Method}
\label{s3}

\subsection{Overview}

In Figure \ref{pipeline}, we present the overall framework of our SDDNet, which adopts an encoder-decoder architecture. During training, SDDNet generates the shadow map, background image, and reconstructed input image; however, only the shadow map is predicted during the inference stage. The generation of reconstructed and background images constitutes our joint training strategy with the main aim of improving the quality of feature disentanglement.

To elaborate, we initially input the image into the backbone network to extract multi-level features $\left\{F_k\right\}_{k=1}^N$, where $N$ represents the number of levels. To fully exploit the detail and global semantics, we divide the features into two groups: the low-level group $F_{low}=\left\{F_k\right\}_{k=1}^{N_{low}}$ and the high-level group ${F_{high}=\left\{F_k\right\}}_{k=N_{low}+1}^N$. We process these two groups in two paths with the same structure, omitting the subscripts $low$ and $high$ for simplicity. The features in each group are concatenated together after upsampling to unify the spatial sizes, generating merged features $\hat{F}$. Subsequently, the FSR module is fed $\hat{F}$ and outputs the shadow-related component ${\hat{F}}^{sd}$, the background-related component ${\hat{F}}^{bg}$, and the recombined features ${\hat{F}}^{re}$. The SSF module then extracts style attributes from ${\hat{F}}^{sd}$, ${\hat{F}}^{bg}$, and ${\hat{F}}^{re}$, and constrains the consistency or diversity of specific style attribute pairs to guide the upstream feature separation. Finally, in the parallel decoder, the shadow-related component, background-related component, and recombined features from both paths are fused separately, and then generate the shadow map, the background image, and the reconstructed input image.

\subsection{Feature Separation and Recombination Module}
\label{fsr}

From the perspective of the human visual system, shadow images can be considered as shadows of other objects superimposed on the background image. This kind of dual-layer separation is not a difficult task for humans, but it is not a simple matter for computers. Therefore, we aim to emulate this way of perceiving shadow images in a bio-inspired manner, and thereby achieve the disentanglement of shadow image content/features. Effective feature disentanglement can promote the focus on more informative components for shadow detection.

One of the main challenges lies in the complex coupling between background and shadow images, making it extremely difficult to model the relationship with complete accuracy. To simplify this process, we model it as a straightforward linear model, which is also a relatively intuitive modeling approach. Nevertheless, to achieve accurate disentanglement within this simple linear model, we design comprehensive strategies based on differentiated supervision. However, differentiated supervision presents its own challenge. Specifically, we only have ground truth shadow maps and lack labels for shadow-free background images, which means that we cannot accomplish our goal solely through direct supervision. Instead, we must find ingenious indirect supervision methods. To this end, in addition to shadow image supervision, we also incorporate joint supervision and style supervision (introduced in the following section). In this manner, the generation of background images can be supervised indirectly, thereby improving the overall feature disentanglement process.

To accomplish this objective, we design the FSR module to achieve feature disentanglement and reorganization through a shadow branch and a background branch. Each branch consists of a residual block \cite{he2016deep}, which comprises two convolutional layers and a skip connection. Given $\hat{F}$, the shadow branch produces the shadow-related component ${\hat{F}}^{sd}$, and the background branch generates the background-related component ${\hat{F}}^{bg}$ as follows:
\begin{equation}
{\hat{F}}^{sd}=Conv\left(Conv\left(\hat{F}\right)\right)+\hat{F},
\end{equation}
\begin{equation}
{\hat{F}}^{bg}=Conv\left(Conv\left(\hat{F}\right)\right)+\hat{F},
\end{equation}
where $Conv$ denotes a convolutional layer. Additionally, we combine them to obtain recombined features ${\hat{F}}^{re}$:
\begin{equation}
{\hat{F}}^{re}={\hat{F}}^{sd}\oplus{\hat{F}}^{bg},
\end{equation}
where $\oplus$ signifies element-wise addition.

Upon obtaining the outputs of the FSR modules in the low- and high-level paths, ${\hat{F}}_{low}^{sd}$, ${\hat{F}}_{low}^{bg}$, ${\hat{F}}_{low}^{re}$ and ${\hat{F}}_{high}^{sd}$, ${\hat{F}}_{high}^{bg}$, ${\hat{F}}_{high}^{re}$ (with subscripts restored), they are individually fused in the parallel decoder:
\begin{equation}
F^{sd}=Conv\left(CA\left(concat({\hat{F}}_{low}^{sd},\ {\hat{F}}_{high}^{sd})\right)\right),
\end{equation}
\begin{equation}
F^{bg}=Conv\left(CA\left(concat({\hat{F}}_{low}^{bg},\ {\hat{F}}_{high}^{bg})\right)\right),
\end{equation}
\begin{equation}
F^{re}=Conv\left(CA\left(concat({\hat{F}}_{low}^{re},\ {\hat{F}}_{high}^{re})\right)\right),
\end{equation}
where $concat$ represents a concatenation operation, and $CA$ denotes channel attention \cite{hu2018squeeze}. By applying channel attention, the network can automatically select informative channels from both low-level and high-level features while suppressing non-informative channels. Finally, the $F^{sd}$, $F^{bg}$, and $F^{re}$ are use to generate the shadow map $P^{sd}$, the background image $P^{bg}$, and the reconstructed input image $P^{re}$, respectively, after passing through a convolutional layer. These processes can be expressed as:
\begin{equation}
P^{sd}=Conv\left(F^{sd}\right),
\end{equation}
\begin{equation}
P^{bg}=Conv\left(F^{bg}\right),
\end{equation}
\begin{equation}
P^{re}=Conv\left(F^{re}\right).
\end{equation}

Although the process for generating these three outputs does not involve distinct operations tailored to their specific targets, we can apply differentiated supervision to enable the network to autonomously learn the optimal way to decompose features. In particular, the supervision for $P^{sd}$ is provided by the ground truth shadow map, while the supervision for $P^{re}$ is derived from the input image. The two losses can be calculated as follows:
\begin{equation}
\mathcal{L}_{sd}=BBCE\left(P^{sd},\ G^{sd}\right),
\end{equation}
\begin{equation}
\mathcal{L}_{re}=MAE(P^{re},\ I),
\end{equation}
where $G^{sd}$ represents the ground-truth shadow map, $I$ denotes the input image, and $BBCE$ and $MAE$ signify the balanced binary cross entropy and mean absolute error, respectively. Here, we employ the same balanced binary cross entropy as in \cite{zhu2021mitigating}, formulated by:
\begin{equation}
\begin{aligned}
BBCE(P^{sd}&, G^{sd}) = \\
&-\sum_{i}
\left[ 
\frac{N_n}{N} G^{sd} log(P_i^{sd}) + \frac{N_p}{N} (1-G^{sd}) log(1-P_i^{sd})
\right],
\end{aligned}
\end{equation}
where $i$ denotes the index of spatial locations, $N_p$ and $N_n$ represent the number of shadow and non-shadow pixels, and $N$ corresponds to the total number of pixels. The mean absolute error is given by:
\begin{equation}
MAE(P^{re},I)=\frac{1}{N}\sum_{i}|P^{re}_i-I_i|.
\end{equation}

The supervision for these two outputs is relatively straightforward. However, for $P^{bg}$, the problem becomes more complex due to the absence of a ground-truth background image. As a result, we use some indirect manners to guide the network learning. In areas without shadows, the input image and the background image are identical, enabling us to directly use the input image to supervise these regions. This can be expressed as:
\begin{equation}
\mathcal{L}_{bg}=MAE\left(P^{bg}\otimes\left(1-P^{sd}\right),\ I\otimes\left(1-G^{sd}\right)\right),
\end{equation}
where $\otimes$ denotes element-wise multiplication. The two terms in this loss correspond to the ground-truth shadow-free regions of the input image and the predicted shadow-free regions of the generated background image. As we employ the predicted shadow-free map $1-P^{sd}$, $\mathcal{L}_{bg}$ has the advantage of constraining the generation of the background image while simultaneously aiding the prediction of the shadow map. Through $\mathcal{L}_{bg}$, we offer guidance to the network for predicting the shadow-free region of the background image. Nevertheless, to predict the complete background image and thereby enhance the quality of disentangling the background-related component, we also need to provide guidance for the shadowed areas. This aspect is accomplished through our SSF module, which will be discussed in Section \ref{ssf}.

\subsection{Shadow Style Filter Module}
\label{ssf}

\begin{figure}[t]
    \centering
    \includegraphics[width=0.48\textwidth]{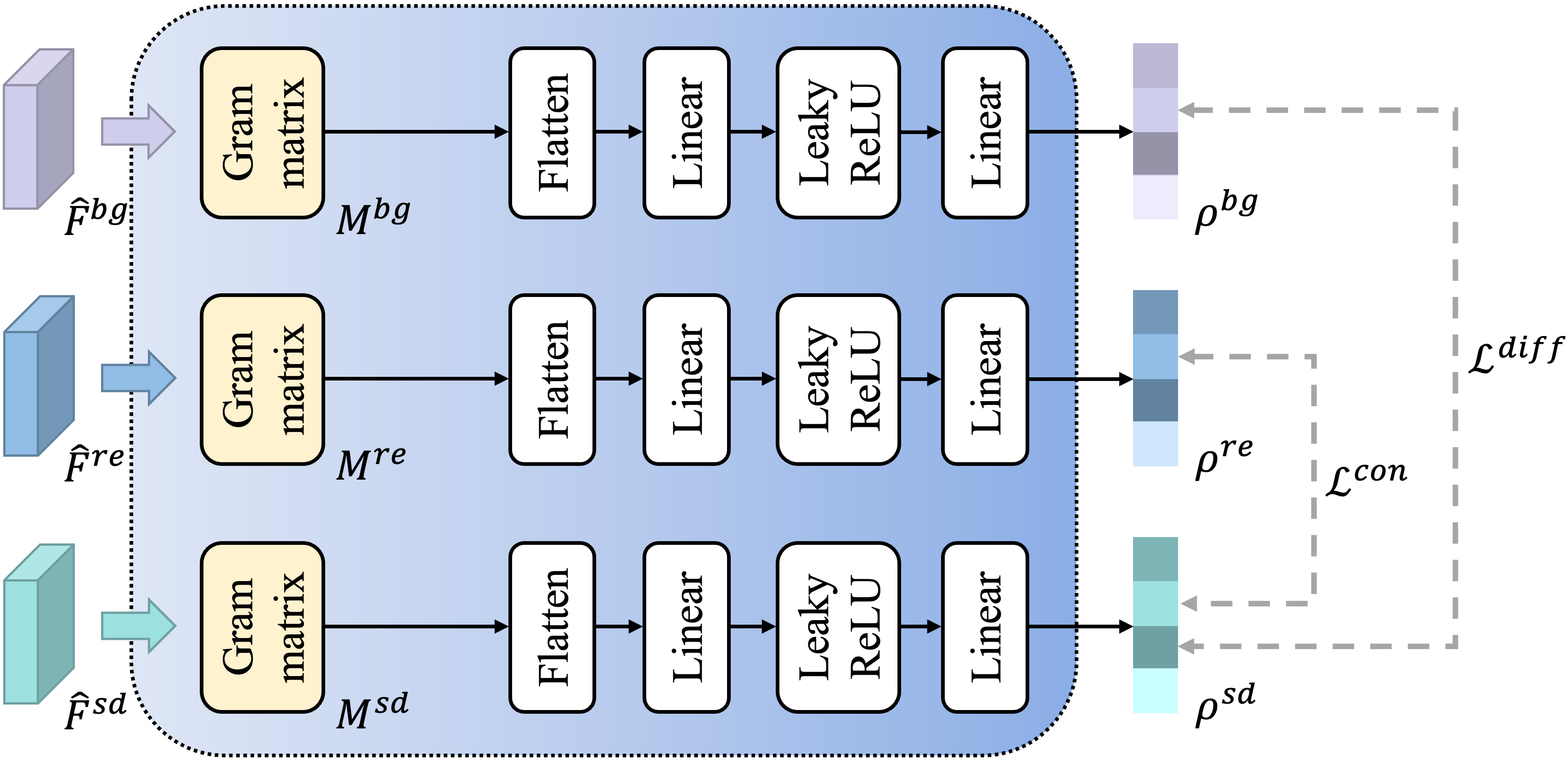}
    \caption{Structure of the SSF module. The Gram matrix is used to extract style attributes of the background-related component, the shadow-related component, and the recombined features. Based on the presence or absence of shadows, we aim to bring the style of the shadow-related component closer to that of the recombined features, while differentiating it with that of the background-related component.}
    \label{ssf:figure}
\end{figure}

In Section \ref{fsr}, we decompose the integrated features into background-related and shadow-related components using the proposed FSR module with a differentiated supervision strategy. However, there are two imperfections: 1) The supervision of the background-related component is insufficient, as it only involves shadow-free regions, leading to a lack of guidance for generating shadowed regions. 2) It does not further emphasize the differences between the shadows and the background, which may results in unclear boundaries for isolating different components, making them less pure.

To address these issues, we consider incorporating style guidance into our method, as the presence or absence of shadows inherently represents a common style attribute. Following this idea, we design the SSF module, as depicted in Figure \ref{ssf:figure}. It extracts style attributes from each of the three outputs from the FSR module  (\ie, ${\hat{F}}^{sd}$, ${\hat{F}}^{bg}$, and ${\hat{F}}^{re}$), and then constrains the consistency and diversity between different style pairs in a contrastive learning fashion. 

For the style attribute extraction, we adopt the Gram matrix \cite{gatys2016image} of the feature map as the style representation. For the input features $\hat{F}\in\mathbb{R}^{C\times H\times W}$, the Gram matrix $M\in\mathbb{R}^{C\times C}$ captures correlations between its channels, which can be computed as follows:
\begin{equation}
M_{x,y}={{\hat{F}}_x}^T{\hat{F}}_y,
\end{equation}
where $M_{x,y}$ denotes the $(x,y)$ element of Gram matrix $M$, and ${\hat{F}}_x$ and ${\hat{F}}y$ represent the $x^{th}$ and $y^{th}$ channels of $\hat{F}$, respectively. Subsequently, we employ two consecutive linear layers to further extract the style attribute $\rho\in\mathbb{R}^{C^2}$ from $M_{x,y}$:
\begin{equation}
\rho=Linear(Linear\left(Flatten\left(M\right)\right)),
\end{equation}
where $Linear$ signifies a linear layer, and $Flatten$ indicates a flatten operation. For the input components and features, ${\hat{F}}^{sd}$, ${\hat{F}}^{bg}$, and ${\hat{F}}^{re}$, the extracted style attribute vectors are denoted as $\rho^{sd}$, $\rho^{bg}$, and $\rho^{re}$, respectively.

In our approach, the primary style consideration is the presence or absence of shadows. From this perspective, the style of the shadow-related component and recombined features should be consistent, as they collectively represent the existence of shadows. To achieve this, we employ the following loss function to enhance their consistency:
\begin{equation}
    \mathcal{L}^{con}=1-cos\left(\rho^{sd},\rho^{re}\right)=1-\frac{\rho^{sd}\cdot\rho^{re}}{\left |\rho^{sd}\right | \left |\rho^{re}\right | },
\end{equation}
in which $cos$ denotes the cosine similarity. Reducing $\mathcal{L}^{con}$ is equivalent to increasing the cosine similarity, which in turn improves the consistency between $\rho^{sd}$ and $\rho^{re}$.

Conversely, the styles of the shadow-related component and background-related component ought to be distinct, as the latter embodies a shadow-free style. To augment their difference, we employ the subsequent differentiate loss:
\begin{equation}
\mathcal{L}^{diff}=\frac{(\rho^{re}\cdot\rho^{bg})^2}{C^2},
\end{equation}
A smaller $\mathcal{L}^{diff}$ signifies that the two vectors are more orthogonal, meaning the difference between them is larger.

The comprehensive style constraint loss, denoted as $\mathcal{L}_{style}$, encompasses the similarity and diversity losses from both low-level and high-level pathways. This loss can be computed using the following equation:
\begin{equation}
\mathcal{L}_{style}=\mathcal{L}_{low}^{con}+\mathcal{L}_{low}^{diff}+\mathcal{L}_{high}^{con}+\mathcal{L}_{high}^{diff}.
\end{equation}

The two constraints in the SSF module enable the two linear layers to extract the style related to the presence or absence of shadows from the Gram matrix more effectively. As the presence or absence of shadows serves as the decisive factor for the diversity or consistency of the two style attribute pairs, if the linear layers were to focus on other styles, the diversity or consistency would not be adequately captured. Thus, the process of back-propagation encourages the linear layers to concentrate on the shadow aspect. With this premise, the constraint that differentiates background-related and shadow-related components fosters the formation of distinctly different characteristics between them. This ensures that the information they contain is not easily duplicated, supporting the feasibility of our dual-layer modeling approach. More importantly, when combined with the shadow-free region constraint described in Section \ref{fsr}, the network gains the ability to separate background component without requiring a ground-truth background image, which in turn refines the shadow-related component.

\subsection{Overall Loss Function}
The overall loss function of our method is formulated as follows:
\begin{equation}
\mathcal{L}=\mathcal{L}_{sd}+\alpha(\mathcal{L}_{re}+\mathcal{L}_{bg})+\beta\mathcal{L}_{style},
\end{equation}
where $\alpha$ and $\beta$ are two balancing hyperparameters, which are empirically set to $\alpha=0.2$ and $\beta=0.1$, respectively.

\section{Experiments}

\subsection{Datasets and Evaluation Metric}

\begin{table*}[t]
  \caption{Quantitative comparison results between our SDDNet and existing state-of-the-art methods. ``Shad." and ``No Shad." denote the error rates of shadow and non-shadow regions, respectively. \textbf{Bold} indicates the best performances, and \underline{underline} indicates the second best performances.}
  \label{QuanComparison}
  \centering
  \begin{tabular}{p{2.7cm}<{\centering}p{1.6cm}<{\centering}p{0.95cm}<{\centering}p{0.95cm}<{\centering}p{1.28cm}<{\centering}p{0.95cm}<{\centering}p{0.95cm}<{\centering}p{1.28cm}<{\centering}p{0.95cm}<{\centering}p{0.95cm}<{\centering}p{1.28cm}<{\centering}}
    \toprule
    \multirow{2}*{Model} & \multirow{2}*{Source} & \multicolumn{3}{c}{ISTD \cite{wang2018stacked}} & \multicolumn{3}{c}{SBU \cite{vicente2016large}} & \multicolumn{3}{c}{UCF \cite{zhu2010learning}} \\
    \cmidrule(r){3-5}
    \cmidrule(r){6-8}
    \cmidrule(r){9-11}
    & & BER$\downarrow$ & Shad.$\downarrow$ & No Shad.$\downarrow$ & BER$\downarrow$ & Shad.$\downarrow$ & No Shad.$\downarrow$ & BER$\downarrow$ & Shad.$\downarrow$ & No Shad.$\downarrow$ \\
    \midrule
    Unary-Pariwise \cite{guo2011single} & CVPR'11 & - & - & - & 25.03 & 36.26 & 13.80 & - & - & - \\
    scGAN \cite{nguyen2017shadow} & ICCV'17 & 4.70 & 3.22 & 6.18 & 9.10 & 8.39 & 9.69 & 11.50 & 7.74 & 15.30 \\
    ST-CGAN \cite{wang2018stacked} & CVPR'18 & 3.85 & 2.14 & 5.55 & 8.14 & 3.75 & 12.53 & 11.23 & 4.94 & 17.52 \\
    DC-DSPF \cite{wang2018densely} & IJCAI'18 & - & - & - & 4.00 & 4.70 & 5.10 & 7.90 & 6.50 & 9.30 \\
    A+D Net \cite{le2018a+} & ECCV'18 & - & - & - & 5.37 & 4.45 & 6.30 & 9.25 & 8.37 & 10.14 \\
    BDRAR \cite{zhu2018bidirectional} & ECCV'18 & 2.69 & 0.50 & 4.87 & 3.64 & 3.40 & 3.89 & 7.81 & 9.69 & 5.44 \\
    DSDNet \cite{zheng2019distraction} & CVPR'19 & 2.17 & 1.36 & 2.98 & 3.45 & 3.33 & 3.58 & 7.59 & 9.74 & 5.44 \\
    DSC \cite{hu2019direction} & TPAMI'19 & 3.42 & 3.85 & 3.00 & 5.59 & 9.76 & 1.42 & 10.54 & 18.08 & 3.00 \\
    MTMT-Net \cite{chen2020multi} & CVPR'20 & 1.72 & 1.36 & 2.08 & 3.15 & 3.73 & 2.57 & 7.47 & 10.31 & 4.63 \\
    ECA \cite{fang2021robust} & MM'21 & 2.03 & 2.88 & 1.19 & 5.93 & 10.82 & 1.03 & 10.71 & 18.59 & 2.83 \\
    RCMPNet \cite{liao2021shadow} & MM'21 & 1.61 & 1.22 & 2.00 & 2.98 & 3.26 & 2.69 & 6.75 & 8.36 & 5.75 \\
    FDRNet \cite{zhu2021mitigating} & ICCV'21 & 1.55 & 1.22 & 1.88 & 3.04 & 2.91 & 3.18 & 7.28 & 8.31 & 6.26 \\
    CM-Net \cite{zhu2022single} & MM'22 & \underline{1.44} & - & - & \textbf{2.94} & - & - & \underline{6.73} & - & - \\
    TransShadow \cite{jie2022fast} & ICASSP'22 & 1.73 & - & - & 3.17 & - & - & 6.95 & - & - \\
    R2D \cite{valanarasu2023fine} & WACV'23 & 1.69 & 0.59 & 2.79 & 3.15 & 2.74 & 3.56 & 6.96 & 8.32 & 5.60 \\
    Ours & / & \textbf{1.27} & 1.01 & 1.52 & \textbf{2.94} & 3.23 & 2.64 & \textbf{6.59} & 7.89 & 5.29 \\
    \bottomrule
  \end{tabular}
\end{table*}

\subsubsection{Datasets}
We evaluate our method on three public datasets: SBU \cite{vicente2016large}, ISTD \cite{wang2018stacked}, and UCF \cite{zhu2010learning}. The SBU dataset comprises 4,089 training images and 638 testing images. The ISTD dataset contains 1,330 training images and 540 testing images. Although it provides ground truths for both shadow maps and shadow-free images, we only use the ground truths for shadow maps in our task. The UCF dataset consists of 135 training images and 110 testing images. Following previous shadow detection works \cite{zhu2021mitigating, zheng2019distraction, chen2020multi, liao2021shadow, hu2019direction, zhu2018bidirectional}, we evaluate our method on both the SBU and UCF test sets using the model trained on the SBU training images, and on the ISTD testing set using the model trained on its own training set.

\subsubsection{Evaluation Metrics}
We following previous shadow detection works \cite{zhu2021mitigating, zhu2022single, nguyen2017shadow} to adopt the widely-used metric, balanced error rate (BER), to quantitatively evaluate performance:
\begin{equation}
BER=\left(1-\frac{1}{2}\left(\frac{TP}{TP+FN}+\frac{TN}{TN+FP}\right)\right)\times100,
\end{equation}
where $TP$, $TN$, $FP$, and $FN$ represent the numbers of true positive, true negative, false positive, and false negative pixels, respectively. BER considers error rates for both shadow and non-shadow regions, with lower values indicating better performance. Additionally, we also report the error rate for the shadow region, $1-\frac{TP}{TP+FN}$, and the error rate for the non-shadow region, $1-\frac{TN}{TN+FP}$.

\begin{figure*}[t]
    \centering
    \includegraphics[width=0.96\textwidth]{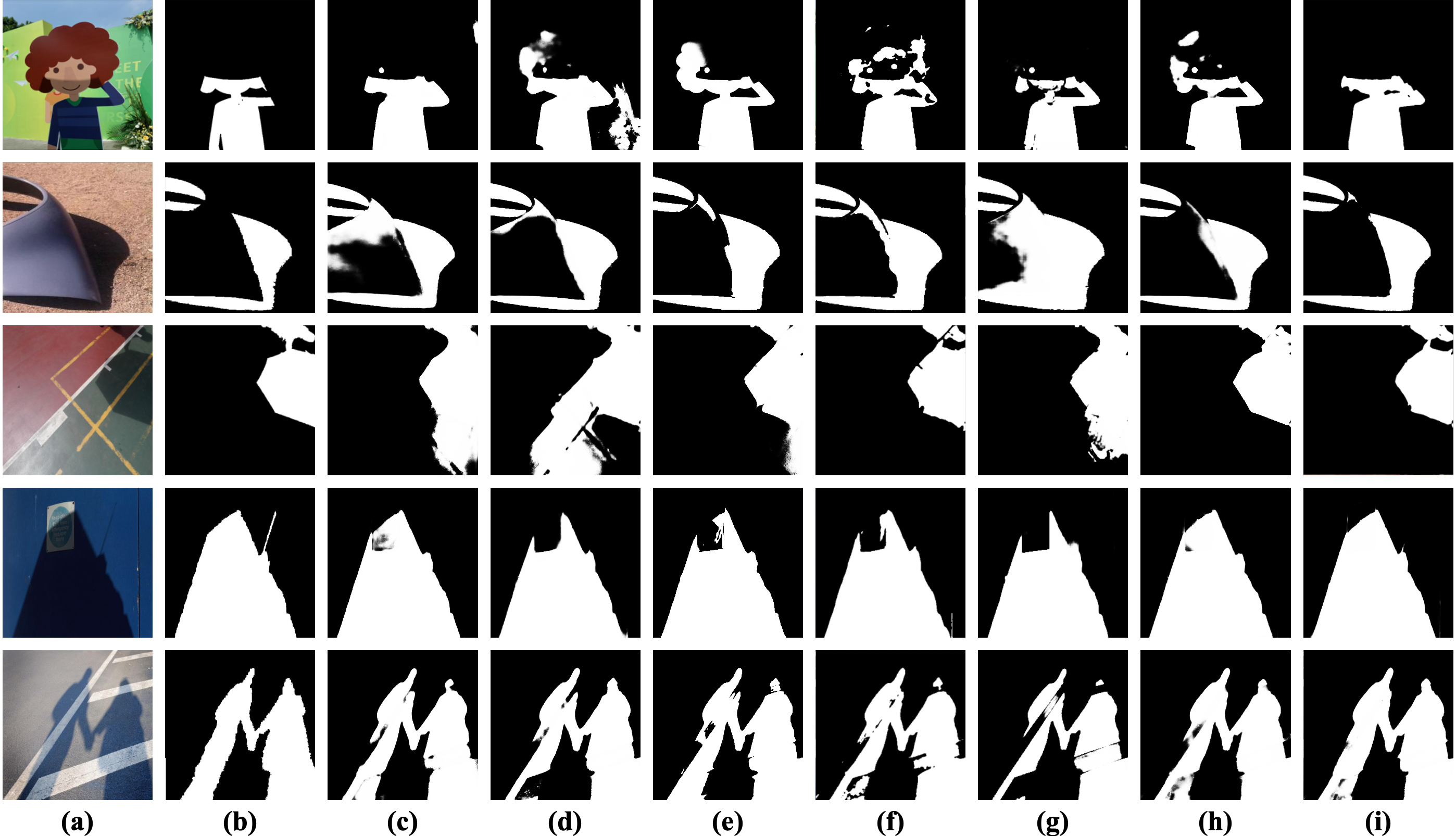}
    \caption{Qualitative comparison between our SDDNet and existing state-of-the-art methods. (a) Input images. (b) Ground-truths. (c) The prediction of BDRAR \cite{zhu2018bidirectional}. (d) The prediction of DSDNet \cite{zheng2019distraction}. (e) The prediction of MTMT-Net \cite{chen2020multi}. (f) The prediction of FDRNet \cite{zhu2021mitigating}. (g) The prediction of ECA \cite{fang2021robust}. (h) The prediction of CM-Net \cite{zhu2022single}. (i) The prediction of our SDDNet.}
    \label{qualitative}
\end{figure*}

\subsection{Implemention Details}
For the backbone, we adopt the lightweight EfficientNet-B3 \cite{tan2019efficientnet} as in \cite{zhu2021mitigating, zhu2022single} and initialize it with pre-trained parameters from ImageNet \cite{deng2009imagenet}. EfficientNet-B3 comprises $25$ consecutive blocks, with the output of the first $6$ layers serving as our low-level features and the output of the remaining layers as the high-level features. Our code is implemented with PyTorch and accelerated by a single NVIDIA RTX2080Ti. We also implement our network by using the MindSpore Lite tool\footnote{\url{https://www.mindspore.cn/}}.

During both training and inference, we resize the input image to $512\times 512$. In the training stage, we optimize the entire network for $20$ epochs with a batch size of $4$, using the Adam optimizer. The initial learning rate is set to $0.0005$. The learning rate is adjusted using the exponential decay strategy, with a decay rate of $0.7$. During the testing stage, we employ a fully connected conditional random field (CRF) \cite{krahenbuhl2011efficient} to further refine our predicted shadow map, following the approaches in \cite{zhu2021mitigating, zheng2019distraction, chen2020multi, liao2021shadow, hu2019direction, zhu2018bidirectional}. 
Our proposed model has a real-time inference speed of 32 FPS for processing an image with the size of $512\times 512$.

\subsection{Comparisons}
We compare our method with 15 previous state-of-the-art shadow detection methods both quantitatively and qualitatively, the methods we select include Unary-Pariwise \cite{guo2011single}, scGAN \cite{nguyen2017shadow}, ST-CGAN \cite{wang2018stacked}, DC-DSPF \cite{wang2018densely}, A+D Net \cite{le2018a+}, BDRAR \cite{zhu2018bidirectional}, DSDNet \cite{zheng2019distraction}, DSC \cite{hu2019direction}, MTMTNet \cite{chen2020multi}, ECA \cite{fang2021robust}, RCMPNet \cite{liao2021shadow}, FDRNet \cite{zhu2021mitigating}, CM-Net \cite{zhu2022single}, TransShadow \cite{jie2022fast}, and R2D \cite{valanarasu2023fine}. Among them, Unary-Pariwise is based on hand-crafted features, while all the others are CNN-based methods. For a fair comparison, all the results are provided directly by the authors or generated by the source codes under the default parameter settings in the corresponding models.

\subsubsection{Quantitative Comparison}

We present the quantitative comparison results between our SDDNet and other models in Table \ref{QuanComparison}. It is clear that our model is highly competitive among all these methods, securing either the first place or a tie for first place in terms of BER across all three datasets. This achievement demonstrates our model's ability to handle data with diverse characteristics and deliver satisfactory outcomes. In comparison to the previously best-performing CM-Net \cite{zhu2022single}, our model exhibits an equal BER on the SBU dataset, while outperforming it by 11.81\% and 2.08\% on the ISTD and UCF datasets, respectively. Additionally, in the comparison of error rates within shadow and non-shadow regions, our model exhibits a consistently stable performance, ranking among the top positions across all three datasets. 

\subsubsection{Qualitative Comparison}

We also qualitatively compare the results of our model with those of previous models, as illustrated in Figure \ref{qualitative}. It can be observed that the results of our model exhibit advantages, particularly in scenes with confusing background colors. For instance, in the first example, the dark eyes and hair of the cartoon character might be misclassified as shadows by other models; however, our SDDNet can effectively mitigate this interference due to its dual-layer modeling. Likewise, in the second and third examples, dark objects or dark ground may be erroneously identified as shadows by other networks, whereas our model prevents this error from occurring.
Moreover, in the fourth and fifth examples, other models may misidentify shadows on light-colored backgrounds as non-shadows due to the varying background colors covered by the shadows. In contrast, our model avoids this interference as the shadow-related component utilized for prediction do not incorporate any background information. 

\subsection{Ablation Study}

To verify the effect of each part in our model, we conduct ablation studies on the SBU dataset with the following configurations:
\begin{itemize}
\item \textit{Baseline}: Compared with the full model introduced in Section \ref{s3}, we remove the FSR module and the SSF module.
\item \textit{Baseline+FSR}: Compared with \textit{Baseline}, we add the FSR module.
\item \textit{Baseline+FSR*}: Compared with \textit{Baseline+FSR}, we remove the joint training of generating the background image and reconstructing the input image, namely remove $\mathcal{L}_{joint}$.
\item \textit{Baseline+FSR+SSF}: Compared with \textit{Baseline+FSR}, we add the SSF module.
\end{itemize}
The quantitative results with all these different configurations are reported in Table \ref{abla}. We also present the quantitative results for several different configurations in Figure \ref{abla_qual}.

\subsubsection{Effectiveness of the FSR module}
In this part, we showcase the effectiveness of our proposed FSR module by comparing its performance to the results obtained without its implementation. The FSR module allows for independent modeling of shadow and background layers, efficiently reducing the adverse effects of confounding background colors. By comparing the performance of \textit{Baseline} and \textit{Baseline+FSR}, it is evident that the FSR module improves the prediction accuracy. Compared to the scenario without the FSR module, the BER score improves from 3.39 to 3.29, with the percentage gain of 2.9\%. 
As illustrated in Figure \ref{abla_qual}, when the backgrounds in shadowed areas (the first example) or non-shadowed areas (the second example) display various distinct characteristics, \textit{Baseline} struggles to eliminate such interference. It predicts the light yellow line in shadows as non-shadow and misclassifies the non-shadow dark part of the red clay court as shadows. In contrast, \textit{Baseline+FSR} performs better, as its isolated shadow-related components can mitigate the impact of background colors to some extent, achieving improved predictions. However, there are still noticeable discrepancies between the result and the ground truth, indicating that the absence of clear guidance for disentangling background-related components hinders the feature disentanglement from achieving complete success.

\begin{table}[!t]
  \caption{Ablation study results for our SDDNet. \textbf{Bold} indicates the best performances.}
  \label{abla}
  \centering
  \begin{tabular}{p{2.6cm}<{\centering}|p{1cm}<{\centering}p{1cm}<{\centering}p{1cm}<{\centering}|p{1cm}<{\centering}}
    \toprule
    Configuration & FSR & $\mathcal{L}_{joint}$ & SSF & BER$\downarrow$ \\
    \cmidrule(r){0-4}
    \textit{Baseline}  &  &  &  & 3.39 \\
    \textit{Baseline+FSR} & \checkmark & \checkmark &  & 3.29 \\
    \textit{Baseline+FSR*}  & \checkmark &  &  & 3.32 \\
    \textit{Baseline+FSR+SSF}  & \checkmark & \checkmark & \checkmark & \textbf{2.94} \\
    \bottomrule
  \end{tabular}
\end{table}

Additionally, we conduct experiments focusing on the joint training strategy in the FSR module, specifically the $\mathcal{L}_{joint}$ term in the loss function. This joint training serves two purposes. Firstly, it constrains the reconstruction of the input image, ensuring that the information within the two isolated components is neither omitted nor redundant. Secondly, it constrains the generation of the background image, encouraging the production of the background-related component, thereby more effectively eliminating the interference of background information from shadow-related components. Compared with \textit{Baseline+FSR} and \textit{Baseline+FSR*}, \textit{Baseline+FSR} that incorporates joint training yields superior results, with a BER improvement of 0.03. This observation demonstrates the significance of joint training, and both of its functions are essential for achieving high-quality feature disentanglement.

\subsubsection{Effectiveness of the SSF module}
Furthermore, we compare the performance of our model with and without the proposed SSF module. This module constrains feature disentanglement, taking into account both style diversity and consistency. By examining the results of \textit{Baseline+FSR} and \textit{Baseline+FSR+SSF} in Table \ref{abla}, it is evident that adding the SSF module yields considerably improved results, with a 0.35 higher BER. This suggests that the style constraints within the SSF module indeed enhance the ability to more effectively separate the two feature components, thus simplifying the prediction of shadow maps. In the absence of the SSF module, disentangling background-related component proves challenging, as ground truth background images are unavailable. However, the SSF module ingeniously addresses this issue by diversifying the styles of background features and shadow-related components in a weakly supervised manner.

\begin{figure}[t]
    \centering
    \includegraphics[width=0.48\textwidth]{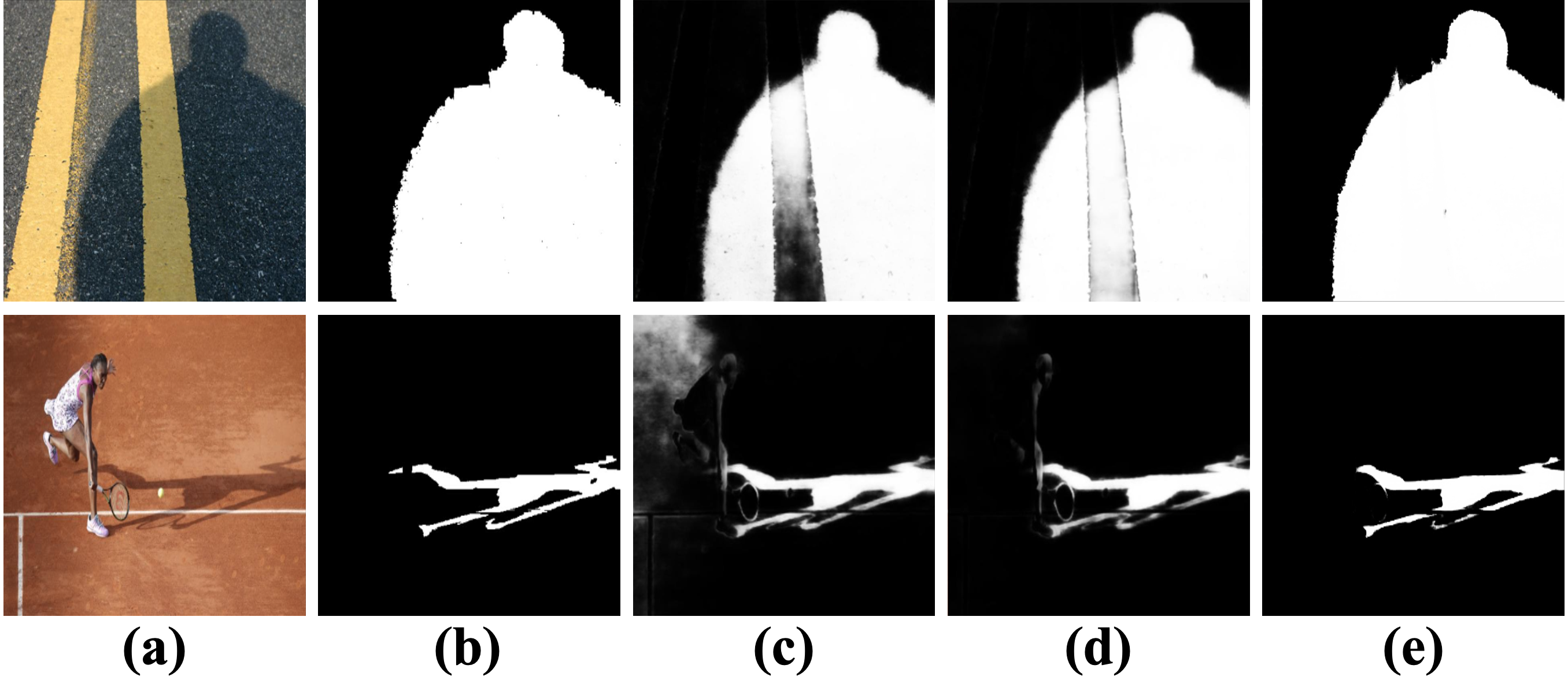}
    \caption{The qualitative results of the ablation study. (a) Input images. (b) Ground-truths. (c) The prediction of \textit{Baseline}. (d) The prediction of \textit{Baseline+FSR}. (e) The prediction of \textit{Baseline+FSR+SSF}.}
    \label{abla_qual}
\end{figure}

In Figure \ref{abla_qual}, we can also observe the superiority brought by the SSF module. The predictions of \textit{Baseline+FSR+SSF} demonstrate an obvious advantage over \textit{Baseline+FSR}, exhibiting a clear improvement in handling complex backgrounds and fully mitigating the impact of confounding background colors. Consequently, both the FSR and SSF modules are indispensable for obtaining stable and robust prediction results. They need to coordinate with each other in order to maximize their effectiveness.

\section{Conclusion}
In this paper, we present a novel Style-guided Dual-layer Disentanglement Network (SDDNet) for shadow detection. Our central idea is to separate the shadow and background layers of the input image to reduce the impact of background color. To achieve this goal, we introduce two novel modules. The first one is the Feature Separation and Recombination (FSR) module that separates complete features into shadow-related and background-related components using differentiated supervisions. Simultaneously, the joint training strategy of reconstructing the input image and generating the background image ensures the reliability of the separation process. Furthermore, we consider the presence and absence of shadows as a type of style and introduce style constraints to our model through a Shadow Style Filter (SSF) module, further enhancing the quality of feature disentanglement. Experimental results on three datasets demonstrate that our SDDNet achieves state-of-the-art performance, proving the effectiveness of our approach.

\begin{acks}
This work was supported in part by the National Key R\&D Program of China under Grant 2021ZD0112100, in part by the National Natural Science Foundation of China under Grant 62002014, Grant U1913204, Grant U1936212, Grant 62120106009, in part by the Taishan Scholar Project of Shandong Province under Grant tsqn202306079, in part by the Project for Self-Developed Innovation Team of Jinan City under Grant 2021GXRC038, in part by the Hong Kong Innovation and Technology Commission (InnoHK Project CIMDA), in part by the Hong Kong GRF-RGC General Research Fund under Grant 11203820 (9042598), in part by Young Elite Scientist Sponsorship Program by the China Association for Science and Technology under Grant 2020QNRC001, and in part by CAAI-Huawei MindSpore Open Fund.
\end{acks}

\bibliographystyle{ACM-Reference-Format}
\balance
\bibliography{ref}










\end{document}